\documentclass[runningheads]{llncs}
\usepackage{amssymb}
\usepackage{graphicx}
\usepackage{subcaption}

\begin{document}

\title{CoinRun: Solving Goal Misgeneralisation}

\author{Stuart Armstrong\inst{1} \and Alexandre Maranh\~{a}o\inst{2}
Oliver Daniels-Koch\inst{3} \and Patrick Leask\inst{4} \and Rebecca Gorman\inst{1}}

\institute{Aligned AI Limited
\and
Instituto Tecnol\'{o}gico de Aeron\'{a}utica (ITA) and CentraleSup\'{e}lec
\and
Charles River Analytics
\and
Durham University}

\authorrunning{Armstrong et al. 2023}


\maketitle

\begin{abstract}
Goal misgeneralisation is a key challenge in AI alignment -- the task of getting powerful Artificial Intelligences to align their goals with human intentions and human morality. In this paper, we show how the ACE (Algorithm for Concept Extrapolation) agent can solve one of the key standard challenges in goal misgeneralisation: the CoinRun challenge. It uses no new reward information in the new environment. This points to how autonomous agents could be trusted to act in human interests, even in novel and critical situations.
\end{abstract}
\section{Introduction}
Artificially intelligent (AI) algorithms are becoming more and more powerful and more and more ubiquitous, shaping the modern world -- and shaping its future. Indeed, it is likely that future AI will have great power over humanity's future \cite{nick2014superintelligence} \cite{russell2019human}. Thus, aligning AI -- getting it to act in humanity's interests \cite{soares2014aligning} -- might be the most important task of this century.

One key challenge in AI alignment is goal misgeneralisation \cite{di2022goal}. It is easy to point an AI towards `proxy goals' that are correlated to the desired goals: we can prime a self-driving car with images of pedestrians, we can train a tank classifier with images of tanks and empty forests\footnote{\url{https://neil.fraser.name/writing/tank/}}, we can instruct an LLM with examples of helpful and harmless responses. And then the car will crush a pedestrian who happened to jay-walk\footnote{\url{https://www.nbcnews.com/tech/tech-news/self-driving-uber-car-hit-killed-woman-did-not-recognize-n1079281}}, the classifier will focus on the weather rather than the tank\footnote{\url{https://buildaligned.ai/blog/concept-extrapolation-for-hypothesis-generation}}, and the LLM will be so helpful that it will provide instructions on how to hotwire a car or threaten to break someone's legs\footnote{\url{https://guzey.com/ai/two-sentence-universal-jailbreak/}}. The problem is that the training data does not provide enough details to nail down the real intended goal; and so the AI selects the simplest goal compatible with the data \cite{shah2020pitfalls}, even if that simple goal is not at all what we intended.

In this paper we show that the goal generalisation problem is solveable. It is possible to get an agent with an underdefined goal, and that can extend this goal intelligently and correctly to new situations. \emph{It does not get or reward information within the new situation}. This is an important step towards eventually solving the alignment problem.

The agent's performance is demonstrated on a simplified goal misgeneralisation example: the CoinRun misgeneralisation problem \cite{di2022goal}.

\section{Related Works}

\textbf{Model Splintering} \cite{splintering} is when conditions of environments change to an extent that the definitions and concepts that were once valid cease to be. Model splintering between training and test environments result in a distribution shift. 

\textbf{Distributions shift} is a difference in the probability distributions between the training and test environments. It can deteriorate the performance of Machine Learning algorithms, leading to out-of-distribution (OOD) failures \cite{10.5555/1462129}\cite{MORENOTORRES2012521}. Robustness to these shifts poses a concern to AI Safety \cite{concrete_problems_ai_safety}\cite{https://doi.org/10.48550/arxiv.2109.13916}\cite{ai_safety_gridworlds}, because at first it is unpredictable how agents will behave in an OOD situation . For high-stakes applications, such as autonomous vehicles \cite{10.1109/CVPR.2019.00498} and healthcare \cite{Castro2020}, the ability to generalize out-of-distribution is critical for deployment \cite{shen_survey}. 

There are dozens of algorithms for OOD generalization and different problem settings that are related to it, such as domain generalization \cite{wang2022generalizing} or domain adaptation \cite{zhang2021survey}. However, none of them eliminate the problem entirely. Gulrajani and Lopez-Paz \cite{gulrajani2020search} have shown that the real effects of various generalization approaches hardly beat the baseline on image data. Even with all the proposed techniques, designing an algorithm with robust OOD generalization performance is still pointed as a prevalent research focus \cite{liu2023outofdistribution}. 

\textbf{Goal Misgeneralization} is a particular form of OOD failure in the setting of reinforcement learning. It happens when an agent is trained to maximize a reward $R$ and then deployed in an environment that is OOD. In the new environment, it maintains its capabilities, i.e., keep acting in a goal-directed manner, but achieves low reward because it seems to pursue a goal different from the specified. It happens when there is a proxy $R' \neq R$ that correlates with the intended objective on the training environment but comes apart in the test environment \cite{di2022goal}. The mismatch between the objectives an agent pursue and those it was trained for is also called an \textit{inner alignment problem} \cite{hubinger2021risks}.

\textbf{Concept extrapolation} \cite{franklin2023concept} is the process of taking features and concepts an agent has learned in training and extending them safely to new environments. Being able to concept extrapolate between environments makes agents robust to model splintering and solve problems such as goal misgeneralization and more general OOD settings. 

\textbf{Diversification} approaches to out-of-distribution generalization try to tackle the problem by training multiple models and then choosing the most appropriate. It is different from model ensembles because these aim to combine multiple models for inference, while diversification approaches train a collection of models and identify one for inference. Teney et al. \cite{teney2022evading} propose to train a collection of models with a diversity regularizer that penalizes aligned gradients to make them rely on different features. DivDis \cite{lee2023diversify} trains a single model with multiple heads, and penalizes mutual information between the predictions in an additional unlabeled dataset. However, to our best knowledge, no work has applied diversification approaches to a RL setting before.

\textbf{Algorithm for Concept Extrapolation (ACE)} is a proprietary algorithm that can use  unlabeled data and diversification to extrapolate concepts to new environments. We used ACE to solve goal misgeneralization, and we see it as a strong candidate to beat OOD generalization benchmarks.

\section{Learning Multiple Reward Hypotheses}
\newcommand{\Acts}{\mathcal{A}}
\newcommand{\States}{\mathcal{S}}
\newcommand{\Envs}{\mathcal{E}}
\subsection{Goal Misgeneralisation}

We will use the formalism of \cite{di2022goal}.

A deep RL agent is trained to maximize a reward $R: \States\times\Acts \to \mathbb{R}$, where $\States$ and $\Acts$ are the sets of all valid states
and actions, respectively. An environment $e$ is a collection of state-action pairs, along with a transition function $T:(\States\times\Acts)^n\to\Delta(\States)$, where $(\States\times\Acts)^n$ is the agent's history (the previous states and actions) and $\Delta(\States)$ is a probability distribution over the next state.

Let $\mathcal{E}$ be a set of environments, with a prior $p_{\Envs}$; for notational convenience, we'll write $\mathcal{E}$ for the $(\mathcal{E},p_{\mathcal{E}})$, suppressing the prior unless explicitly noted.

Then define $\pi^R|_{\Envs}$ as a policy that maximises the expected reward\footnote{The existence of this policy requires some assumptions on $R$ and $\Envs$ to ensure convergence; in this paper, the sets of states and actions are finite, which is sufficient.} of $R$ in the environments set $\Envs$.

Assume that the agent is deployed out-of-distribution; that is, an aspect of the environment (and therefore the distribution of observations) changes at test time. Goal misgeneralisation occurs if the agent now achieves low reward in the new environment because it continues to act capably yet appears to optimize a different reward. Formally:
\begin{definition}
Let $R$ be the true reward, and let $\Envs$ be a set of (training) environments and $\Envs'$ a set of (out-of-distribution, test) environments. An agent undergoes goal misgeneralisation if there exists a reward function $R'$ and the agent follows policy $\pi$ with:
\begin{eqnarray*}
    \pi |_{\Envs} & \approx \pi^{R'} |_{\Envs} \approx \pi^R |_{\Envs} \\
    \pi |_{\Envs'} & \approx \pi^{R'} |_{\Envs'} \not\approx \pi^R |_{\Envs'}.
\end{eqnarray*}
\end{definition}
In other words, the agent behaves as an $R'$-maximiser in both sets of environments, but only in $\Envs$ is this also the behaviour of an $R$-maximiser. We call $R$ the intended objective and $R'$ the behavioral objective of the agent.



\subsection{CoinRun misgeneralisation}

\begin{figure}
    \centering
    \includegraphics[width=0.8\textwidth]{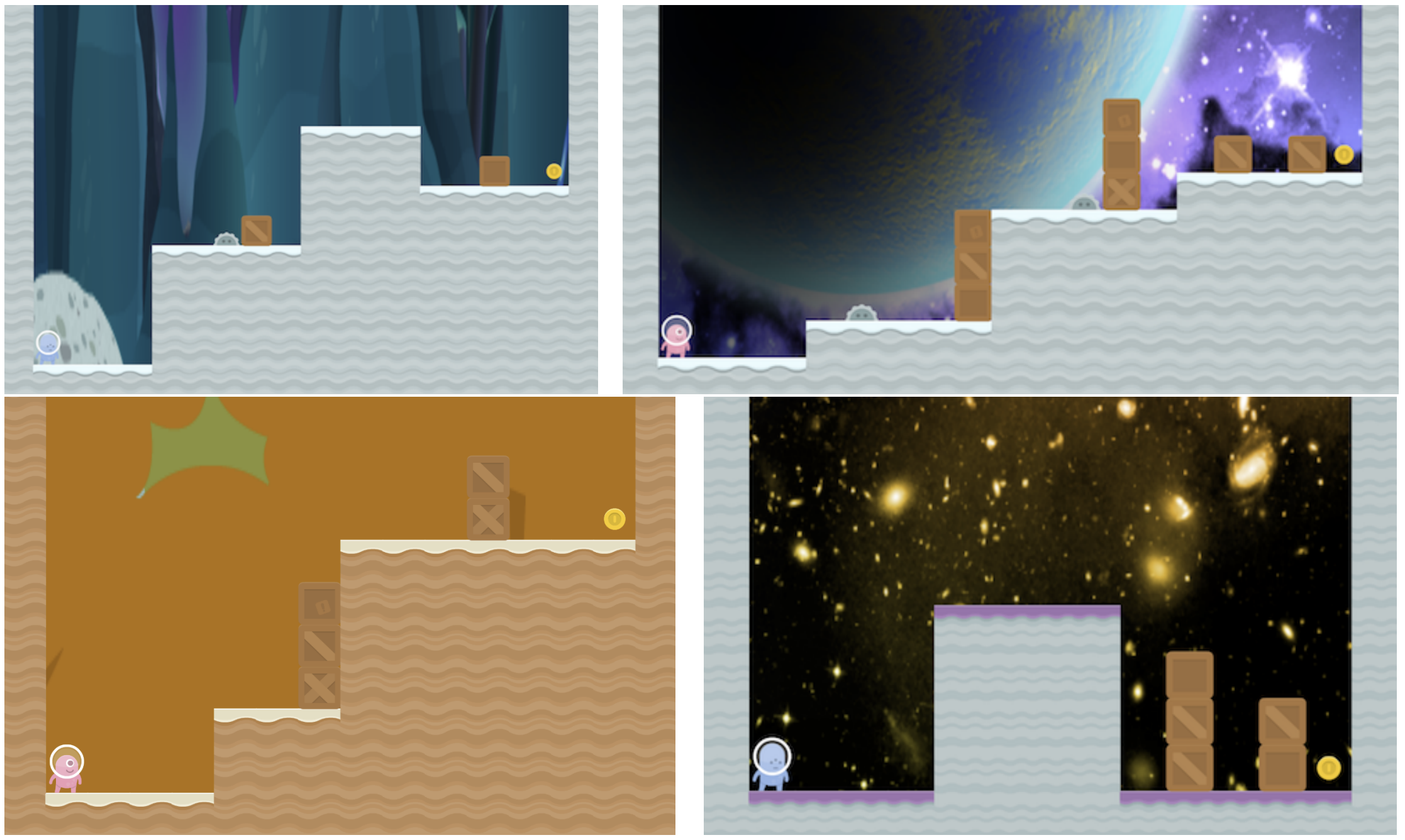}
    \caption{Four generated CoinRun levels with the coin on the right.}
    \label{coinrun:examples}
\end{figure}

CoinRun \cite{cobbe2019quantifying} is a RL problem created to evaluate the generalization performance of trained agents. It consists on levels where the agent spawns on the far left and a single coin (see Figure \ref{coinrun:examples}). The objective is to collect the coin, but there are several obstacles both stationary (`lava') and non-stationary (`monsters'). A collision with an obstacle results in the level's termination; it also ends after $1,000$ timesteps. The only reward is obtained by collecting the coin.

The true reward $R$ is thus $1$ when the agent first gets the coin, and $0$ otherwise.

When testing for goal misgeneralisation, $\Envs$ is a set of environments where the coin is always placed at the right end of the level. The behavioral objective, as we'll see, is $R'$ which gives $1$ when the agent first gets to the right of the level. Obviously, on $\Envs$, $R=R'$ (getting to the right is the same as getting the coin), thus $\pi^R|_{\Envs}=\pi^{R'}|_{\Envs}$.

\begin{figure}
    \centering
    \includegraphics[width=0.8\textwidth]{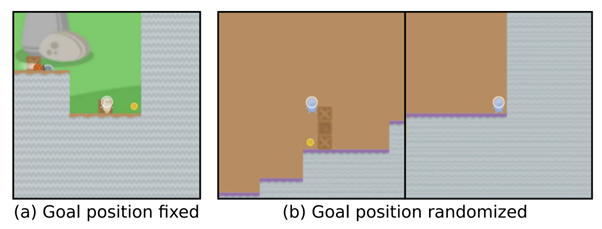}
    \caption{In a), the $R$=`get the coin' and the $R'$=`go to the right' reward functions are equivalent; levels like these are used as training data, where the agent will receive the true reward. In b), the rewards are not equivalent; in these levels, the agent will never know or receive a reward.}
    \label{coinrun:ood}
\end{figure}

Then on the testing environments, $\Envs'$, the position of the coin was randomised (see Figure \ref{coinrun:ood}). Here, $R$ and $R'$ are now distinct rewards. \emph{On $\Envs'$, the agents will not receive any reward information},

\subsubsection{Standard agent: goal misgeneralisation}

Following \cite{di2022goal}, we trained an agent on $\Envs$; agents were trained on $100K$ procedurally generated levels for $200M$ timesteps.

Then they were tested on $50K$ different levels drawn from $\Envs'$. The agent successfully got the coin in $59.13\%$ of the levels.

Does this mean that the agent avoid goal misgeneralisation? Does it get the coin "in passing", or does it aim for the coin specifically\footnote{Note that this score is quite different from that in \cite{di2022goal} (Figure 2); an author of that paper has confirmed that they used a different baseline, but that our scoring method is reasonable.}

To check for that, we first need a decent baseline agent to compare to. The baseline agent randomly cycles between the following actions: move right, move right-up (jump right) and move right-down. The baseline agent got the coin $55.50\%$ of the time. It is plausible that the $+3.63\%$ overperformance of the standard trained agent is just due to it learning to avoid monsters, lava, and getting stuck.

\begin{figure}
    \centering
    \includegraphics[width=0.4\textwidth]{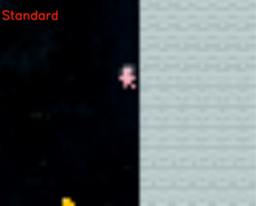}
    \caption{The standard agent is stuck jumping against the right wall, ignoring the coin visible on the bottom behind it.}
    \label{agent:stuck}
\end{figure}

To test that this is indeed the case, we studied the behaviour of the standard agent. Indeed, it didn't seek out the coin, and continued moving right past it. In Figure \ref{agent:stuck}, the trained agent continues to jump up and down forever, ignoring the coin to its left. So it's clear that it's following the reward function $R'$, rather than the true reward $R$.

\subsubsection{ACE agent: correct goal generalisation}

The ACE (`Algorithm for Concept Extrapolation') agent was trained on the training environment $\Envs$, and continued to learn on the testing environment $\Envs'$, \emph{though it never got any reward information in those environments}.

\begin{figure}
    \centering
    \includegraphics[width=0.8\textwidth]{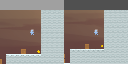}
    \caption{The basic data that the ACE agent uses.}
    \label{double:image}
\end{figure}

The agent initially trained in $\Envs$, getting information about situations that provided rewards versus situations that didn't. Because the agent has ``momentum'', it uses the last two images, of the two previous states. The bar on the top encodes the action that the agent took, with different shades of grey corresponding to different actions (see Figure \ref{double:image}).

It then explored $50$ different testing environments $\Envs'$, moving mainly right, and saving the images history of the first $50$ time-steps (it never got any reward information here, either explicitly or implicitly). It then used the (model-agnostic) ACE algorithm to analyse the new data and compare with the known reward and non-rewards in $\Envs$.

It then produces two hypotheses, $R_0$ and $R_1$, for what the `true' reward function could be. The $R_0$ corresponded to $R'$ and the $R_1$ corresponded to $R$. This allows us to generate a `prudent' agent: one that maximises the average of $R_0$ and $R_1$.

\begin{figure}
\begin{subfigure}{.5\textwidth}
  \centering
  \includegraphics[width=.8\linewidth]{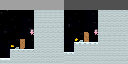}
  \caption{High score on $R_0$.}
  \label{fig:sfig1}
\end{subfigure}%
\begin{subfigure}{.5\textwidth}
  \centering
  \includegraphics[width=.8\linewidth]{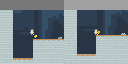}
  \caption{High score on $R_1$.}
  \label{fig:sfig2}
\end{subfigure}
\caption{Indicative images of high scores for the two candidate functions.}
\label{reward:choice}
\end{figure}

Alternatively, the agent could present winning conditions to a human overseer, and get us to choose the correct behaviour. For example, it could present the two images in \ref{reward:choice} to a human overseer. The human would select the second image as indicative of the correct reward function. This single bit of information informs the agent as to which reward is correct\footnote{Ultimately the agent will be able to select this itself, but that is for subsequent papers.} -- namely $R_1$.

Then, with $R_1$ selected, it was then trained for $100K$ timesteps on $\Envs'$, using its own estimated $R_1$ for reward (without any actual reward information from the environment).

\begin{figure}
    \centering
    \includegraphics[width=0.8\textwidth]{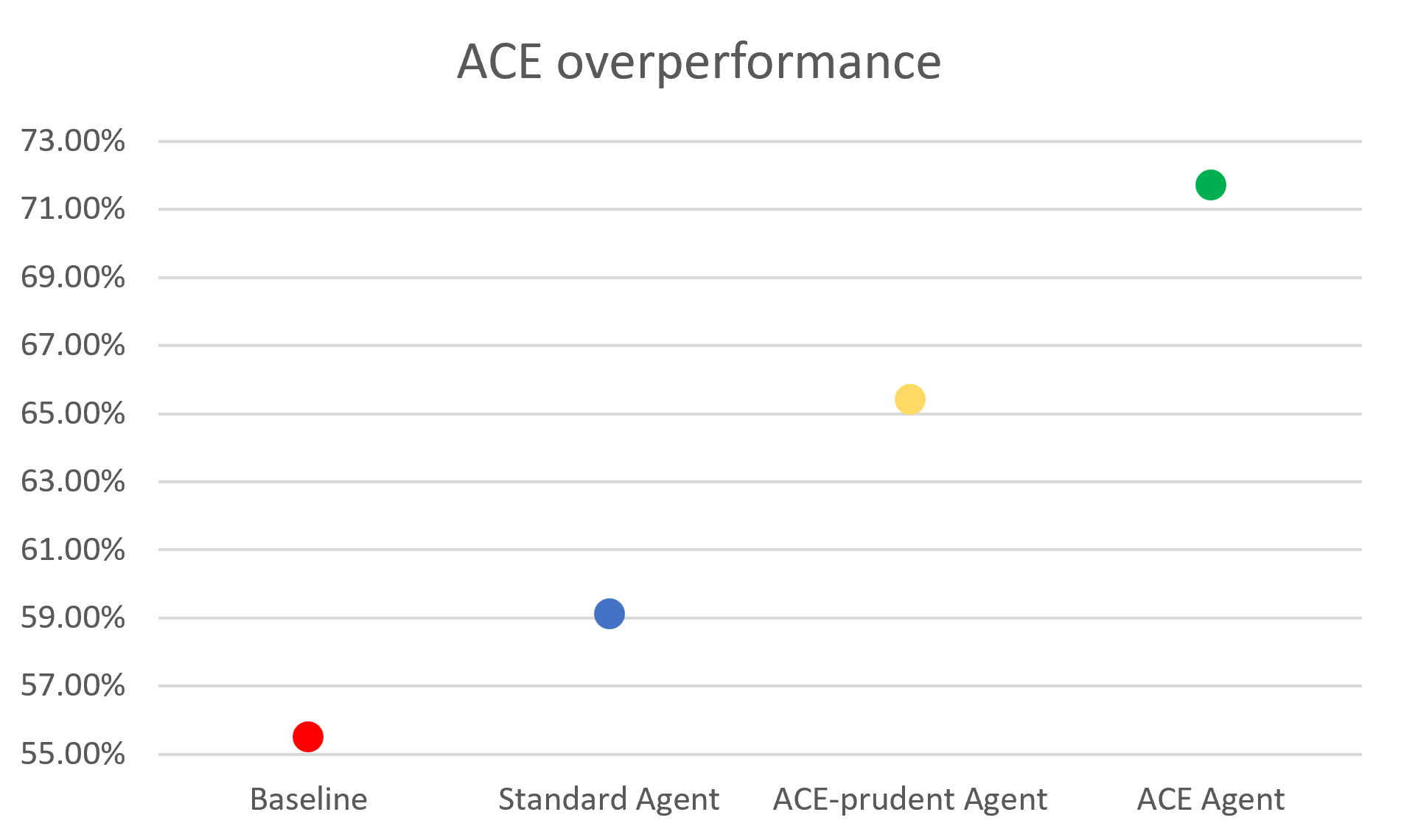}
    \caption{Offline performances of the four agents on random locations of the coin without additional reward information. `Baseline' is a simple policy of 'always go right'. `Standard Agent', trained on labeled training data, goes right while also avoiding monsters and holes, and only picks up the coin by accident on the way. Without additional reward information, `ACE-Prudent Agent' learns to disambiguate getting the coin from going to the right and tries to achieve both goals. The `ACE Agent' presents informative images of two possible reward functions, getting one bit of human feedback on which is correct.}
    \label{fig:performances}
\end{figure}

After this, it was tested on $50K$ different levels in $\Envs'$, and strongly outperforms the standard agent: it got the coin $71.70\%$ of the time; that's a $16.20\%$ improvement over the baseline, a more than four-fold improvement over the standard agent ($3.63\%$). See Figure \ref{fig:performances}. Also plotted is the performance of the `prudent-ACE' agent, which did not get the single bit of extra information and thus maximised the average of $R_0$ and $R_1$. It finds the coin $65.42\%$ of the time, a more than two-fold improvement over the standard agent.

\begin{figure}
    \centering
    \includegraphics[width=0.8\textwidth]{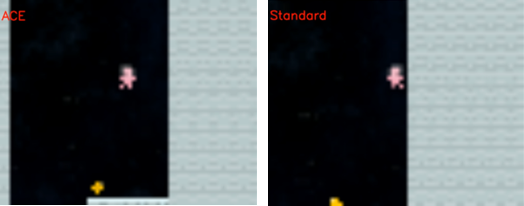}
    \caption{The standard agent is stuck jumping against the right wall, ignoring the coin visible on the bottom behind it.}
    \label{agent:unstuck}
\end{figure}

Importantly, we can check that the ACE agent actually is following $R$ rather than $R'$. For instance, the ACE agent returns to get the coin in the same situation where the standard agent stays stuck; see Figure \ref{agent:unstuck}.

\begin{figure}
    \centering
    \includegraphics[width=0.8\textwidth]{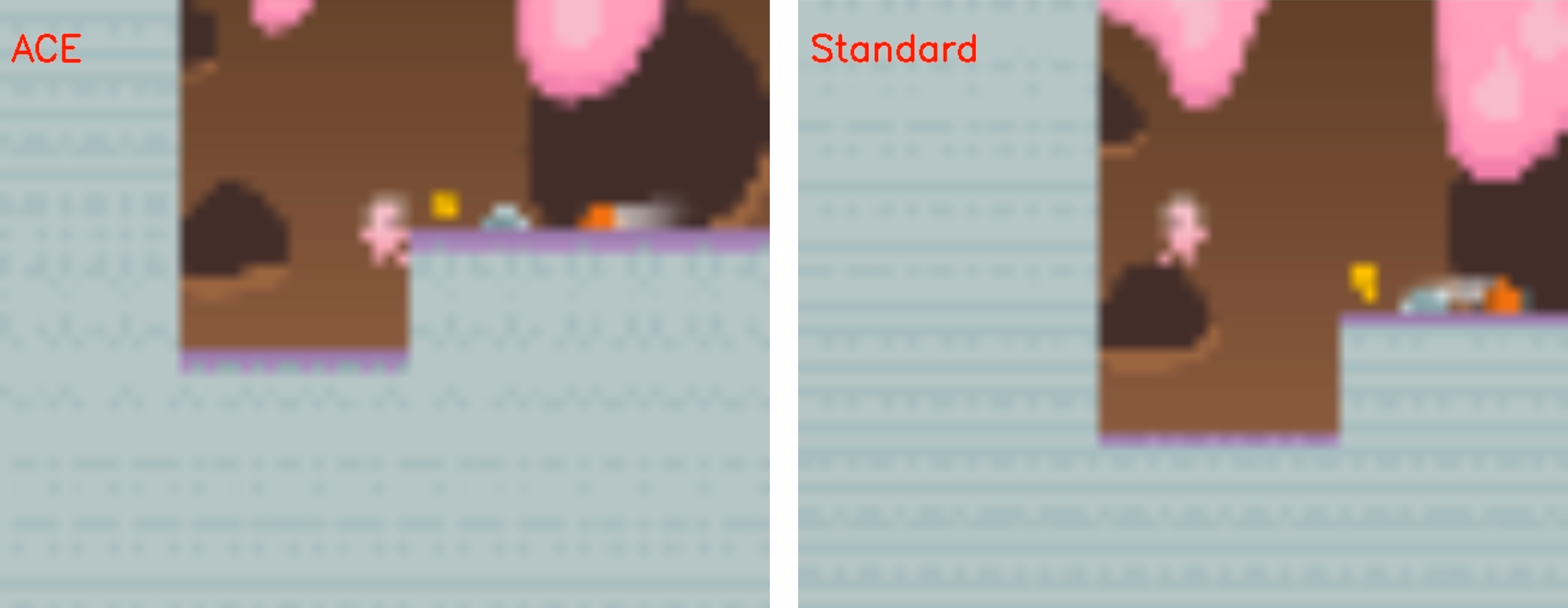}
    \caption{The standard agent doesn't dare go near the coin, because of the monsters, as it wants to get around them and go to the right. The ACE agent goes straight for the coin, winning its reward before the monsters get near.}
    \label{agent:stuck:left}
\end{figure}

There are also situations where the ACE and standard agent's behaviour differ in interesting ways; in the following level, the standard agent stays stuck on the left, too afraid to go near the monsters\footnote{Please pardon the anthropomorphisation.}. But the ACE agent jumps straight to the coin, because it `knows' that the monsters are irrelevant once it's got the reward; see Figure \ref{agent:stuck:left}

\section{Conclusion}

Generalisation in one test environment is a good start. The ACE algorithm, being model-agnostic, should be able to solve many of the goal misgeneralisation test problems, and could scale to a full solution of goal misgeneralisation. Beyond the goal misgeneralisation problem is the overall generalisation problem: how to get an AI to robustly extent its goals and capabilities to any new environment. It is our hope that this paper is just the first step towards that goal.

\bibliographystyle{splncs04}
\bibliography{mybibliography}

\begin{thebibliography}{10}
\providecommand{\url}[1]{\texttt{#1}}
\providecommand{\urlprefix}{URL }
\providecommand{\doi}[1]{https://doi.org/#1}

\bibitem{10.1109/CVPR.2019.00498}
Alcorn, M., Li, Q., Gong, Z., Wang, C., Mai, L., Ku, W.S., Nguyen, A.: Strike (with) a pose: Neural networks are easily fooled by strange poses of familiar objects. pp. 4840--4849 (06 2019). \doi{10.1109/CVPR.2019.00498}

\bibitem{concrete_problems_ai_safety}
Amodei, D., Olah, C., Steinhardt, J., Christiano, P., Schulman, J., Mané, D.: Concrete problems in ai safety (2016). \doi{10.48550/ARXIV.1606.06565}, \url{https://arxiv.org/abs/1606.06565}

\bibitem{splintering}
Armstrong, S.: Model splintering: moving from one imperfect model to another (2020), \url{https://www.lesswrong.com/s/u9uawicHx7Ng7vwxA/p/k54rgSg7GcjtXnMHX}

\bibitem{Castro2020}
Castro, D.C., Walker, I., Glocker, B.: Causality matters in medical imaging. Nature Communications  \textbf{11}(1), ~3673 (Jul 2020). \doi{10.1038/s41467-020-17478-w}, \url{https://doi.org/10.1038/s41467-020-17478-w}

\bibitem{cobbe2019quantifying}
Cobbe, K., Klimov, O., Hesse, C., Kim, T., Schulman, J.: Quantifying generalization in reinforcement learning. In: International Conference on Machine Learning. pp. 1282--1289. PMLR (2019)

\bibitem{di2022goal}
Di~Langosco, L.L., Koch, J., Sharkey, L.D., Pfau, J., Krueger, D.: Goal misgeneralization in deep reinforcement learning. In: International Conference on Machine Learning. pp. 12004--12019. PMLR (2022)

\bibitem{franklin2023concept}
Franklin, M., Gorman, R., Ashton, H., Armstrong, S.: Concept extrapolation: A conceptual primer (2023)

\bibitem{gulrajani2020search}
Gulrajani, I., Lopez-Paz, D.: In search of lost domain generalization (2020)

\bibitem{https://doi.org/10.48550/arxiv.2109.13916}
Hendrycks, D., Carlini, N., Schulman, J., Steinhardt, J.: Unsolved problems in ml safety (2021). \doi{10.48550/ARXIV.2109.13916}, \url{https://arxiv.org/abs/2109.13916}

\bibitem{hubinger2021risks}
Hubinger, E., van Merwijk, C., Mikulik, V., Skalse, J., Garrabrant, S.: Risks from learned optimization in advanced machine learning systems (2021)

\bibitem{lee2023diversify}
Lee, Y., Yao, H., Finn, C.: Diversify and disambiguate: Learning from underspecified data (2023)

\bibitem{ai_safety_gridworlds}
Leike, J., Martic, M., Krakovna, V., Ortega, P.A., Everitt, T., Lefrancq, A., Orseau, L., Legg, S.: Ai safety gridworlds (2017). \doi{10.48550/ARXIV.1711.09883}, \url{https://arxiv.org/abs/1711.09883}

\bibitem{liu2023outofdistribution}
Liu, J., Shen, Z., He, Y., Zhang, X., Xu, R., Yu, H., Cui, P.: Towards out-of-distribution generalization: A survey (2023)

\bibitem{MORENOTORRES2012521}
Moreno-Torres, J.G., Raeder, T., Alaiz-Rodríguez, R., Chawla, N.V., Herrera, F.: A unifying view on dataset shift in classification. Pattern Recognition  \textbf{45}(1),  521--530 (2012). \doi{https://doi.org/10.1016/j.patcog.2011.06.019}, \url{https://www.sciencedirect.com/science/article/pii/S0031320311002901}

\bibitem{nick2014superintelligence}
Nick, B.: Superintelligence: Paths, dangers, strategies  (2014)

\bibitem{10.5555/1462129}
Quionero-Candela, J., Sugiyama, M., Schwaighofer, A., Lawrence, N.D.: Dataset Shift in Machine Learning. The MIT Press (2009)

\bibitem{russell2019human}
Russell, S.: Human compatible: Artificial intelligence and the problem of control. Penguin (2019)

\bibitem{shah2020pitfalls}
Shah, H., Tamuly, K., Raghunathan, A., Jain, P., Netrapalli, P.: The pitfalls of simplicity bias in neural networks. Advances in Neural Information Processing Systems  \textbf{33},  9573--9585 (2020)

\bibitem{shen_survey}
Shen, Z., Liu, J., He, Y., Zhang, X., Xu, R., Yu, H., Cui, P.: Towards out-of-distribution generalization: A survey (2021). \doi{10.48550/ARXIV.2108.13624}, \url{https://arxiv.org/abs/2108.13624}

\bibitem{soares2014aligning}
Soares, N., Fallenstein, B.: Aligning superintelligence with human interests: A technical research agenda. Machine Intelligence Research Institute (MIRI) technical report  \textbf{8} (2014)

\bibitem{teney2022evading}
Teney, D., Abbasnejad, E., Lucey, S., van~den Hengel, A.: Evading the simplicity bias: Training a diverse set of models discovers solutions with superior ood generalization (2022)

\bibitem{wang2022generalizing}
Wang, J., Lan, C., Liu, C., Ouyang, Y., Qin, T., Lu, W., Chen, Y., Zeng, W., Yu, P.S.: Generalizing to unseen domains: A survey on domain generalization (2022)

\bibitem{zhang2021survey}
Zhang, Y.: A survey of unsupervised domain adaptation for visual recognition (2021)

\end{thebibliography}

\end{document}